\documentclass{article}
\usepackage{spconf,amsmath,amssymb,graphicx,hyperref,booktabs,bm,multirow,microtype,cite}
\hypersetup{hidelinks}
\let\oldthebibliography\thebibliography
\renewcommand{\thebibliography}[1]{%
  \oldthebibliography{#1}%
  \fontsize{10pt}{10pt}\selectfont
  \setlength{\itemsep}{4.5pt}
  \setlength{\parskip}{0pt}%
}

\title{S$^3$N: A Spherical Spiral Scanning Network for Weather Forecasting}

\name{%
\begin{tabular}{c}
Fan Yan$^{1}$\qquad
Chen Hui$^{1,\ast}$\qquad
Weisi Lin$^{2}$\quad
Haiqi Zhu$^{3}$\\ 
Feng Jiang$^{1}$\qquad
Sun-Yuan Kung$^{4}$\qquad
Wei Zhang$^{5}$%
\end{tabular}%
\thanks{$^{\ast}$Corresponding author:
\texttt{chui@nuist.edu.cn}.}%
}

\address{%
$^{1}$Nanjing University of Information Science and Technology \quad
$^{2}$Nanyang Technological University\\
$^{3}$Harbin Institute of Technology\quad
$^{4}$Princeton University\quad
$^{5}$Karolinska Institutet
}

\begin{document}

\maketitle

\begin{abstract}
Machine learning–based weather prediction (MLWP) has achieved strong performance in global weather forecasting.\ Recent Hierarchical Equal Area isoLatitude Pixelation (HEALPix)-based methods use the HEALPix (HP) grid to avoid area distortion near the poles of conventional latitude–longitude (LL) grids.\ However, existing HP-based approaches often use pointwise mapping methods and process HP pixels within separate base faces or local windows.\ Consequently, the mapping may introduce reconstruction errors and cross-face communication depends on handcrafted boundary handling or shifted windows.\ We propose the Spherical Spiral Scanning Network (S$^3$N) to address both limitations. First, L2Proj provides a bidirectional method for mapping atmospheric fields between the LL and HP grids through an $L^2$ projection of their continuous finite-element representations. Second, the Attention-Guided Quad-Spiral State-Space Scanning (AQSS) block uses cross-latitude attention to guide selective state-space updates along four global pole-to-pole spiral paths. This design enables continuous information propagation across HP base-face boundaries without additional boundary-processing mechanisms.\ Experiments show that S$^3$N achieves better results at 4-, 7-, and 10-day lead times, and exhibits slower error growth in long-range forecasting.
\end{abstract}

\begin{keywords}
Weather Forecasting, Deep Learning, HEALPix, Mamba, Attention
\end{keywords}

\section{Introduction}
\label{sec:intro}

\begin{figure}[htb]
  \centering
  \centerline{\includegraphics[width=8.5cm]{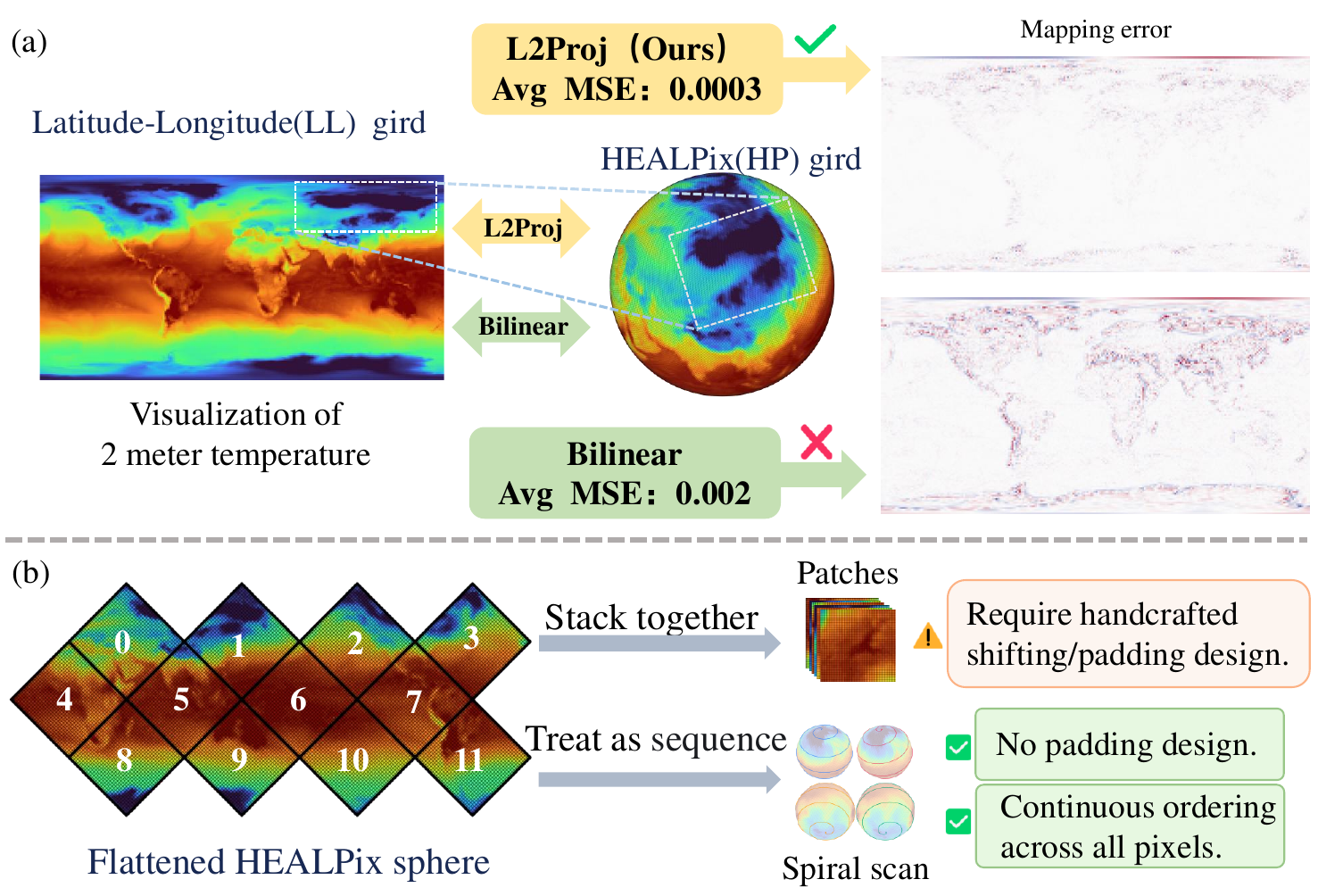}}
  \vspace{-6pt}
\caption{Comparison of various methods. (a) The average error of L2Proj is much lower than that of bilinear interpolation. (b) We treat HEALPix pixels from the twelve base faces as a sequence rather than stacking the twelve base faces as patches, thereby avoiding handcrafted boundary handling.}
\vspace{-19pt}
\label{fig:motivation}
\end{figure}
Machine-learning weather prediction (MLWP) has made great progress in recent years, represented by Pangu-Weather, FourCastNet, GraphCast, NeuralGCM, and Aurora~\cite{bi2023pangu,pathak2022fourcastnet,lam2023graphcast,kochkov2024neuralgcm,bodnar2025aurora}. Most MLWP methods still use the regular latitude--longitude (LL) grid~\cite{chen2023fuxi,chen2025fengwu,nguyen2023climax,nguyen2024stormer,bonev2023sfno,lang2024aifs,price2025gencast}.\ But the physical area represented by each LL grid cell decreases as latitude increases toward the poles, and most methods use latitude-weighted objectives to compensate for the unequal sampling during optimization. However, this only adjusts the training weights and does not change the LL-grid representation, which remains physically nonuniform and oversamples the polar regions.

HEALPix provides an alternative equal-area representation of the sphere~\cite{gorski2005healpix}, which has been adopted by deep learning methods such as DeepSphere, HEALPix CNNs, OSLO, SCE-Depth, and HEALPix-guided diffusion models~\cite{perraudin2019deepsphere,krachmalnicoff2019healpixcnn,bidgoli2022oslo,zhu2026scedepth,bao2026illumination}, and weather forecasting models including DLWP-HPX, HEAL-ViT, ESyM~\cite{karlbauer2024dlwphpx,ramavajjala2024healvit,cresswellclay2025dlesym}.\ However, existing HEALPix-based weather forecasting methods still face two practical limitations. First, their LL--HP mapping usually relies on pointwise or learned resampling, while the two grids differ in sampling density and spatial resolution
\cite{karlbauer2024dlwphpx,ramavajjala2024healvit,linander2025pear}. Second, most methods organize the HP grid into separate patches, which limits the propagation of information across adjacent faces and requires additional mechanisms such as rotated or copied padding, shifted windows, masks, or layout-specific indexing for cross-boundary communication~\cite{karlbauer2024dlwphpx,ramavajjala2024healvit,carlsson2024healswin}.

To address these two limitations, we develop a unified framework that combines LL--HP data mapping with a global cross-face information propagation mechanism.

First, we introduce L2Proj, a bidirectional mapping method based on surface finite-element principles \cite{dziuk1988surfacefem,dziuk2013surfacefem}. L2Proj represents the LL and HP data as continuous piecewise-linear fields on the unit sphere and maps between them through an $L^2$ projection. This provides a unified mapping method for the two nonmatching spherical grids.

Second, we introduce the Attention-Guided Quad-Spiral State-Space Scanning (AQSS) block, inspired by selective state-space models and multidirectional visual scanning
\cite{gu2024mamba,liu2024vmamba}. As shown in Fig.~\ref{fig:motivation}(b), AQSS arranges all HP pixels along four global pole-to-pole spiral paths. Cross-latitude attention guides the state-space updates and the fusion of the four scan outputs. This design enables the propagation of information across HP base-face boundaries without handcrafted padding, shifted windows, or layout-specific boundary operations.

Our main contributions are as follows:
\vspace{-2pt}
\begin{itemize}
\vspace{-2pt}
    \item We present S$^3$N, a spherical weather forecasting framework that combines finite-element-based LL--HP mapping with an attention-guided cross-face state-space scanning mechanism.\ S$^3$N achieves a mean normalized RMSE (nRMSE) of 0.5177 at a 240-hour lead time, compared with 0.6398 for OneForecast~\cite{gao2025oneforecast}, demonstrating slower error growth during long-range forecasting.
\vspace{-2pt}

    \item We introduce L2Proj, a bidirectional mapping method based on surface finite-element principles. It represents LL and HP data as continuous piecewise-linear fields on the unit sphere and maps between them through an $L^2$ projection.  L2Proj achieves lower errors than bilinear interpolation when mapping  atmospheric fields from the LL grid to the HP grid and then back to the LL grid.
\vspace{-2pt}

    \item We propose the Attention-Guided Quad-Spiral State-Space Scanning (AQSS) block, which uses cross-latitude attention to guide selective state-space updates along four global pole-to-pole spiral paths.\ This design allows hidden states to propagate directly across HP base-face boundaries without face-wise padding or shifted-window processing.
\end{itemize}

\vspace{-6pt}
\section{Method}
\label{sec:method}

\begin{figure*}[t]
  \centering
  \includegraphics[width=1 \linewidth]{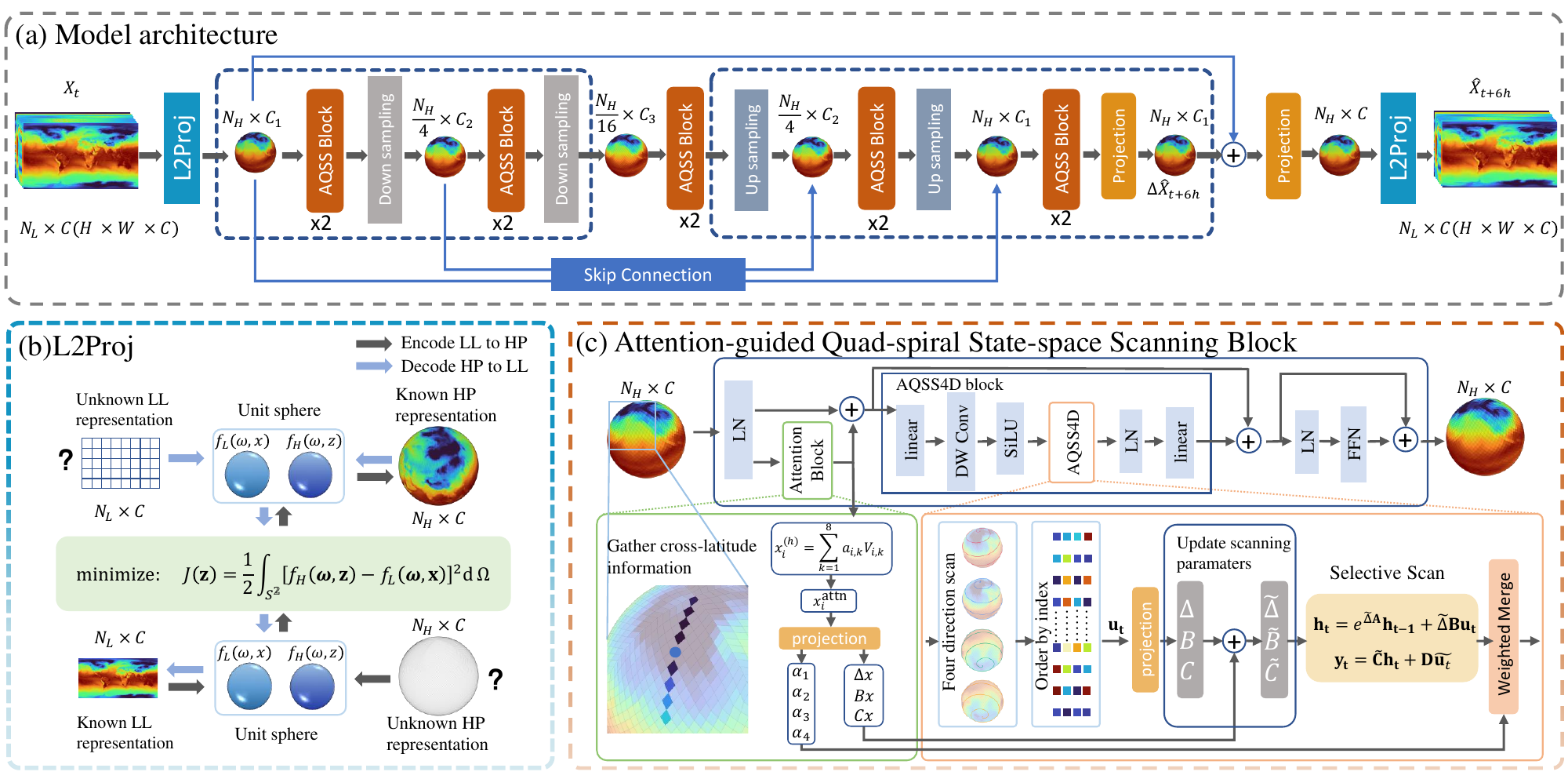}
  \vspace{-0.8cm}
  \caption{Overview of our S$^3$N framework. (a) The overall pipeline of our model, which follows an encoder-processor-decoder architecture. (b) Illustration of the L2Proj computation. We use L2Proj as the encoder and decoder for projecting atmospheric fields between the LL and the HP grids. (c) Details of the proposed AQSS block. Cross-latitude information is collected from eight cross-latitude neighbors to guide the Mamba updates and the weighted fusion of the four path features.}
\vspace{-10pt}
  \label{fig:model_architechture}
\end{figure*}

\subsection{Overview}
\vspace{-2pt}

Our framework, shown in Fig.~\ref{fig:model_architechture}(a), consists of two components: the L2Proj mapping module and the AQSS forecasting backbone. We use L2Proj to map atmospheric fields between the LL grid and the HP grid. The AQSS block receives the input field and performs multiscale spherical feature modeling through four global spiral scans and attention-guided state-space processing. Cross-latitude attention provides spatial context, which is used to guide the Mamba updates and the weighted fusion of the four path features.
\vspace{-8pt}
\subsection{Spherical \texorpdfstring{$L^2$}{L2} Finite-Element Projection}

Given an atmospheric field defined on an LL grid with spatial resolution \(H_L \times W_L\), we use a HEALPix grid with \(12n_{\mathrm{side}}^2\) pixels for its spherical representation. On the LL grid, all longitude points in each polar row correspond to the same physical location at the pole; therefore, each polar row is merged into a single finite-element node. After processing, the LL grid contains $N_L$ nodes while the HEALPix grid contains $N_H$ nodes. The two grids are then mapped onto the unit sphere and triangulated independently.

For any point $\boldsymbol{\omega}$ on the sphere, we find the triangle containing it on each mesh and then compute the weights of the vertices of this triangle using first-order linear interpolation. This allows the discrete values defined at the grid nodes to be extended to a continuous field over the entire sphere. For an atmospheric variable, let $x_i$ denote its value at LL node $i$ and $z_a$ denote its value at HP node $a$. The continuous fields of the LL and the HP grids are defined as:
{
\setlength{\abovedisplayskip}{2pt}
\setlength{\belowdisplayskip}{2pt}
\setlength{\abovedisplayshortskip}{2pt}
\setlength{\belowdisplayshortskip}{2pt}
\begin{equation}
    f_L(\boldsymbol{\omega};\mathbf{x})
    =
    \sum_{i=1}^{N_L}
    x_i\phi_i(\boldsymbol{\omega}),
    \quad
    f_H(\boldsymbol{\omega};\mathbf{z})
    =
    \sum_{a=1}^{N_H}
    z_a\psi_a(\boldsymbol{\omega}),
    \label{eq:continuous_fields}
\end{equation}
where $\phi_i$ and $\psi_a$ are the P1 basis functions associated with LL node $i$ and HP node $a$, respectively.

We determine the HP representation $\mathbf{z}$ by minimizing the global difference between the two continuous fields over the entire unit sphere:
{
\setlength{\abovedisplayskip}{2pt}
\setlength{\belowdisplayskip}{2pt}
\setlength{\abovedisplayshortskip}{2pt}
\setlength{\belowdisplayshortskip}{2pt}
\begin{equation}
J(\mathbf{z})
=
\frac{1}{2}
\int_{\mathbb{S}^{2}}
\left[
f_H(\boldsymbol{\omega};\mathbf{z})
-
f_L(\boldsymbol{\omega};\mathbf{x})
\right]^2
\,\mathrm{d}\Omega,
\end{equation}
}where $\mathbb{S}^{2}$ denotes the unit sphere and $\mathrm{d}\Omega$ denotes its surface-area element. Because the continuous HP field $f_H$ is linear with respect to $\mathbf{z}$, this can be written as the quadratic form:
{
\setlength{\abovedisplayskip}{2pt}
\setlength{\belowdisplayskip}{2pt}
\setlength{\abovedisplayshortskip}{2pt}
\setlength{\belowdisplayshortskip}{2pt}
\begin{equation}
    J(\mathbf{z})
    =
    \frac{1}{2}\mathbf{z}^{\mathsf{T}}M_H\mathbf{z}
    -
    \mathbf{z}^{\mathsf{T}}B\mathbf{x}
    +
    c(\mathbf{x}),
    \label{eq:l2_quadratic}
\end{equation}
}where $c(\mathbf{x})$ is independent of $\mathbf{z}$. The two matrices that determine the optimal HP representation are:
{
\setlength{\abovedisplayskip}{2pt}
\setlength{\belowdisplayskip}{2pt}
\setlength{\abovedisplayshortskip}{2pt}
\setlength{\belowdisplayshortskip}{2pt}
\begin{equation}
    \resizebox{0.91\linewidth}{!}{$ \displaystyle
        (M_H)_{ab}
    =
    \int_{\mathbb{S}^{2}}
    \psi_a(\boldsymbol{\omega})
    \psi_b(\boldsymbol{\omega})
    \,\mathrm{d}\Omega,
    \;
    B_{ai}
    =
    \int_{\mathbb{S}^{2}}
    \psi_a(\boldsymbol{\omega})
    \phi_i(\boldsymbol{\omega})
    \,\mathrm{d}\Omega,
    $}
    \label{eq:fem_matrices}
\end{equation}
}the HP mass matrix $M_H$ encodes the overlap of the HP P1 basis functions over the sphere. The cross-grid coupling matrix $B$ measures the overlap between the LL and HP P1 basis functions over the sphere. Setting the derivative of Eq.~\eqref{eq:l2_quadratic} with respect to $\mathbf{z}$ to zero gives:
{
\setlength{\abovedisplayskip}{2pt}
\setlength{\belowdisplayskip}{2pt}
\setlength{\abovedisplayshortskip}{2pt}
\setlength{\belowdisplayshortskip}{2pt}
\begin{equation}
    M_H\mathbf{z}=B\mathbf{x},
    \label{eq:ll_to_hp_system}
\end{equation}
}$M_H$ and $B$ depend only on the positions of the spherical nodes and the resolutions of the two grids. Therefore, they can be precomputed offline and reused during training and inference, without mesh-intersection processing or numerical integration at runtime. Similarly, the HP-to-LL mapping is constructed independently in the same way.

\vspace{-8pt}
\subsection{Attention-Guided Quad-Spiral Scanning Block}
After L2Proj mapping, the atmospheric state is represented on the HP grid. For one stage, let $\mathbf{X}\in\mathbb{R}^{C\times N}$ denote the input feature, where $C$ is the number of feature channels and
$N=12n_{\mathrm{side}}^2$ is the number of HP pixels. An AQSS block first computes local spherical context and then propagates this information globally through four complete spiral scans.

For each HP pixel ${\mathbf{x}_{i}}$, we sample eight cross-latitude neighbors $\widetilde{\mathbf{x}}_{ik}$, with four located northward and four southward at offsets of one to four HP pixel intervals, as illustrated in Fig.~\ref{fig:model_architechture}(c). The sampling follows the corresponding great circle and remains continuous across a geographic pole. The center feature is then used as the query, while the eight neighbor features are used as keys and values for multi-head attention. For each attention head $h$, the computation is:
{
\setlength{\abovedisplayskip}{2pt}
\setlength{\belowdisplayskip}{2pt}
\setlength{\abovedisplayshortskip}{2pt}
\setlength{\belowdisplayshortskip}{2pt}
\begin{equation}
    \resizebox{0.91\linewidth}{!}{$ \displaystyle
    a_{ik}^{(h)}
    =
    \mathrm{softmax}_{k}
    \left(
    \frac{\mathbf{q}_i^{(h)\mathsf{T}}\mathbf{k}_{ik}^{(h)}}{\sqrt{d}}
    + b_{ik}^{(h)}
    \right),
    \quad
    \mathbf{x}_{i}^{(h)}
    =
    \sum_{k=1}^{8}
    a_{ik}^{(h)}
    \mathbf{v}_{ik}^{(h)},
    $}
    \label{eq:geodesic_attention}
\end{equation}
}where $\mathbf{q}_i^{(h)}=W_Q^{(h)}\mathbf{x}_i$ is the query of pixel $i$, $\mathbf{k}_{ik}^{(h)}=W_K^{(h)}\widetilde{\mathbf{x}}_{ik}$ and $\mathbf{v}_{ik}^{(h)}=W_V^{(h)}\widetilde{\mathbf{x}}_{ik}$ are the key and value of its $k$-th cross-latitude neighbor, while $b_{ik}^{(h)}$ is a learned bias that describes the relative direction and distance between the center pixel and its neighbor on the sphere. The coefficient $a_{ik}^{(h)}$ is the normalized attention weight of neighbor $k$.

\vspace{-2pt}
The outputs of all $H$ attention heads are concatenated and projected to obtain the final context feature:
{
\setlength{\abovedisplayskip}{2pt}
\setlength{\belowdisplayskip}{2pt}
\begin{equation}
\mathbf{x}_{i}^{\mathrm{attn}}
=
W_O
\operatorname{Concat}
\left(
\mathbf{x}_{i}^{(1)},
\ldots,
\mathbf{x}_{i}^{(H)}
\right),
\label{eq:multihead_context}
\end{equation}
}where $\mathbf{x}_i^{\mathrm{attn}}$ is the fused attention context of pixel $i$ from all $H$ heads, and $W_O$ projects the concatenated multi-head features back to the original channel dimension. The context feature is computed before spiral scanning and is used to guide the selective state-space model. The context feature first modulates the scan input through a residual connection:
{
\setlength{\abovedisplayskip}{2pt}
\setlength{\belowdisplayskip}{2pt}
\setlength{\abovedisplayshortskip}{2pt}
\setlength{\belowdisplayshortskip}{2pt}
\begin{equation}
    \mathbf{u}_i
    ={\mathbf{x}}_i
    +
    \lambda_x F_x(\mathbf{x}_{i}^{\mathrm{attn}}),
    \label{eq:context_input_modulation}
\end{equation}
}where $F_x$ is a learned projection and $\lambda_x$ controls the
strength of the context contribution. The cross-latitude context also adjusts the
dynamic Mamba parameters:
{
\setlength{\abovedisplayskip}{2pt}
\setlength{\belowdisplayskip}{2pt}
\setlength{\abovedisplayshortskip}{2pt}
\setlength{\belowdisplayshortskip}{2pt}
\begin{align}
    \widetilde{\Delta}^{(p)}
    &=
    {\Delta}^{(p)}
    +
    \lambda_{\Delta}
    \tanh\!\left(F_{\Delta}(\mathbf{x}_{i}^{\mathrm{attn}})\right),
    \nonumber\\
    \widetilde{\mathbf{B}}^{(p)}
    &=
    {\mathbf{B}}^{(p)}
    +
    \lambda_B
    \tanh\!\left(F_B(\mathbf{x}_{i}^{\mathrm{attn}})\right),
    \nonumber\\
    \widetilde{\mathbf{C}}^{(p)}
    &=
    {\mathbf{C}}^{(p)}
    +
    \lambda_C
    \tanh\!\left(F_C(\mathbf{x}_{i}^{\mathrm{attn}})\right),
    \label{eq:context_ssm_modulation}
\end{align}
}where $\widetilde{\Delta}^{(p)}$, $\widetilde{\mathbf{B}}^{(p)}$, and $\widetilde{\mathbf{C}}^{(p)}$ control state retention, input writing, and state readout, respectively.

\vspace{-2.8pt}
To propagate information globally, we construct four directional spiral paths: $P_{N2S+}$, $P_{N2S-}$, $P_{S2N+}$, and $P_{S2N-}$. The four scan paths are constructed using the HEALPix RING ordering. Specifically, $P_{N2S+}$ and $P_{N2S-}$ scan from the North Pole to the South Pole in clockwise and counterclockwise directions, respectively, while $P_{S2N+}$ and $P_{S2N-}$ scan from the South Pole to the North Pole in clockwise and counterclockwise directions, respectively. Each path visits every HP pixel exactly once, forming a continuous pole-to-pole spiral. The selective scan process is represented as:
{
\setlength{\abovedisplayskip}{2pt}
\setlength{\belowdisplayskip}{2pt}
\setlength{\abovedisplayshortskip}{2pt}
\setlength{\belowdisplayshortskip}{2pt}
\begin{align}
    \mathbf{h}^{(p)}_t
    &=
    e^{\widetilde{\Delta}^{(p)}_t \mathbf{A}}
    \odot
    \mathbf{h}^{(p)}_{t-1}
    +
    \widetilde{\Delta}^{(p)}_t
    \widetilde{\mathbf{B}}^{(p)}_t
    \mathbf{u}^{(p)}_t,
    \nonumber\\
    \mathbf{y}^{(p)}_t
    &=
    \widetilde{\mathbf{C}}^{(p)}_t
    \mathbf{h}^{(p)}_t
    +
    \mathbf{D}\odot\mathbf{u}^{(p)}_t,
\end{align}
}where $p$ indexes the different scan paths, and $t$ indexes the position along the path; $\mathbf{h}^{(p)}_t$ is the corresponding hidden state; $e^{\widetilde{\Delta}^{(p)}_t \mathbf{A}}$ indicates how much of the previous state is retained along the scan path, while $\mathbf{D}$ indicates how much of the current input is passed directly to the output without going through the state. 
\vspace{-1pt}
After scanning, each path output is restored to the original HP pixel order and merged using weights generated from $\mathbf{x}_{i}^{\mathrm{attn}}$, as illustrated in Fig.~\ref{fig:model_architechture}(c).

\vspace{-4pt}
\section{Experiments}
\label{sec:exp}

\subsection{Dataset and Implementation Details}
\label{sec:exp_set}

We use ERA5 reanalysis data~\cite{hersbach2020era5} from WeatherBench
2~\cite{rasp2024weatherbench2}, produced by the European Centre for Medium-Range Weather Forecasts (ECMWF) \cite{molteni1996ecmwf}. The dataset has a spatial resolution of
$1.5^\circ$ ($121\times240$ grid points), a temporal resolution of 6 hours, and 69 atmospheric fields. We use 1979--2017 for training, 2018--2019 for validation, and 2020 for testing. The HP representation uses $n_{\mathrm{side}}=64$ and contains 49,152 equal-area pixels. S$^3$N is trained for one-step prediction followed by long-range autoregressive fine-tuning. Training takes less than 12 days on one NVIDIA RTX 6000D GPU with 84\,GB memory. All comparative experiments are conducted under the same conditions.

\vspace{-6pt}
\subsection{Bidirectional Mapping Evaluation}
\label{sec:mapping_result}
\vspace{-6pt}

{
\setlength{\textfloatsep}{1pt}
\setlength{\floatsep}{3pt}
\setlength{\intextsep}{3pt}
\begin{table}[htbp]
    \centering
    \caption{Mean squared error (MSE) of atmospheric fields after mapping from the LL grid to the HP grid and then back to the LL grid. Lower is better.}
    \label{tab:mapping_mse}
    \setlength{\tabcolsep}{12pt}
    \begin{tabular}{lcc}
        \toprule
        Method & Bilinear Interpolation & L2Proj \\
        \midrule
        Average MSE & 0.002153 & \textbf{0.000252} \\
        \bottomrule
    \end{tabular}
    \vspace{-4pt}
\end{table}
}

We evaluate the mapping performance of L2Proj and bilinear interpolation using data from 2017 and 2020. For both methods, the 69 standardized atmospheric fields are first mapped from the LL grid to the HP grid and then remapped back to the LL grid, where the reconstructed fields are compared with the original LL fields. Table~\ref{tab:mapping_mse} presents the average MSE over all variables after bidirectional mapping. L2Proj achieves a lower reconstruction error.

\vspace{-6pt}
\subsection{Main Results}
\label{sec:main_results}
\vspace{-6pt}

{
\setlength{\textfloatsep}{1pt}
\setlength{\floatsep}{3pt}
\setlength{\intextsep}{3pt}
\begin{table}[htbp]
  \centering
  \caption{Comparison of mean normalized RMSE (lower is better) across different lead times. The best result in each column is in bold, and the second best is underlined.}
  \label{tab:rmse_comparison}
  \resizebox{\linewidth}{!}{%
    \begin{tabular}{lccccc}
      \toprule
      Method & 6-hour & 1-day & 4-day & 7-day & 10-day \\
      \midrule
      Pangu-Weather~\cite{bi2023pangu} & 0.1182 & 0.1859 & 0.4509 & 0.6960 & 0.7582 \\
      FuXi ~\cite{chen2023fuxi} & 0.1258 & 0.2050 & 0.4742 & 0.6484 & 0.7831 \\
      Fengwu ~\cite{chen2025fengwu} & 0.1252 & 0.2028 & 0.4860 & 0.6887 & 0.8143 \\
      GraphCast~\cite{lam2023graphcast} & \underline{0.0784} & \underline{0.1519} & 0.3599 & 0.5521 & 0.7006 \\
      OneForecast~\cite{gao2025oneforecast} & \textbf{0.0754} & \textbf{0.1393} & \underline{0.3275} & \underline{0.5042} & \underline{0.6398} \\
      S$^3$N (Ours) & 0.1009 & 0.1663 & \textbf{0.3171} & \textbf{0.4411} & \textbf{0.5177} \\
      \bottomrule
    \end{tabular}%
  }
  \vspace{-6pt}
\end{table}
}

Table~\ref{tab:rmse_comparison} compares the mean normalized RMSE of different forecasting models across lead times. S$^3$N achieves the best performance at 4-, 7-, and 10-day lead times, with nRMSE values of 0.3171, 0.4411, and 0.5177, respectively. In particular, its advantage becomes more pronounced as the forecast horizon increases, reducing the 10-day nRMSE by 19.1\% compared with the second-best OneForecast. Although S$^3$N is less competitive at short lead times of 6 hours and 1 day, the results demonstrate its stronger capability in maintaining forecast accuracy over longer horizons.

{
\setlength{\textfloatsep}{1pt}
\setlength{\floatsep}{3pt}
\setlength{\intextsep}{3pt}
\begin{table}[htbp]
  \centering
  \vspace{-10pt}
  \caption{Absolute nRMSE growth between adjacent lead times and total growth from 6 hours to 10 days. Lower is better.}
  \label{tab:error_growth}

  \setlength{\tabcolsep}{3pt}
  \small

\begin{tabular}{lccccc}
  \toprule
  \multirow{3}{*}{\raisebox{4pt}{Method}}
  & 6-hour & 1-day & 4-day & 7-day & 6-hour \\[-4pt]

  & to
  & to
  & to
  & to
  & to \\[-4pt]

  & 1-day & 4-day & 7-day & 10-day & 10-day \\
  \midrule

      Pangu-Weather & 0.0677 & 0.2650 & 0.2451 & \textbf{0.0622}& 0.6400 \\
      FuXi & 0.0792 & 0.2692 & \underline{0.1742} & 0.1347 &0.6573 \\
      Fengwu & 0.0776 & 0.2832 & 0.2027 & 0.1256 & 0.6891 \\
      GraphCast & 0.0735 & 0.2080 & 0.1922 & 0.1485 & 0.6222\\
      OneForecast & \textbf{0.0639} & \underline{0.1882} & 0.1767 & 0.1356 &\underline{0.5644} \\
      S$^3$N (Ours) & \underline{0.0654} & \textbf{0.1508} & \textbf{0.1240} & \underline{0.0766} & \textbf{0.4168}\\
    \bottomrule
  \end{tabular}
  \vspace{-12pt}
\end{table}
}

Table~\ref{tab:error_growth} further compares absolute nRMSE growth across forecast horizons. S$^3$N achieves the lowest growth from 1 to 4 days (0.1508), from 4 to 7 days (0.1240), and from 6 hours to 10 days (0.4168). It also ranks second in the remaining intervals, indicating slower error accumulation and stronger long-range forecasting stability.

\vspace{-6pt}
\subsection{Ablation Study}
\label{sec:ablation_study}
\vspace{-6pt}

{
\setlength{\textfloatsep}{1pt}
\setlength{\floatsep}{3pt}
\setlength{\intextsep}{3pt}

\begin{table}[htb]
  \centering
  \caption{Component ablation study of our AQSS. Mean normalized RMSE (lower is better) across different lead times. The best result is in bold, and the second best is underlined. ``w/o'' means ``without'' , ``F. W.'' means ``Fixed-Weight''.}
  \label{tab:ablation}

  \fontsize{9.5pt}{9.5pt}\selectfont
  \setlength{\tabcolsep}{1.2pt}
  \renewcommand{\arraystretch}{1.0}

  \begin{tabular}{@{}lccccc@{}}
    \toprule
    Method & 6-hour & 1-day & 4-day & 7-day & 10-day \\
    \midrule

    S$^3$N (Full)
      & \textbf{0.1009}
      & \textbf{0.1663}
      & \textbf{0.3171}
      & \textbf{0.4411}
      & \textbf{0.5177} \\

    Fixed-Weight Path Fusion
      & 0.2536
      & 0.4428
      & \underline{0.5657}
      & \underline{0.5936}
      & \underline{0.6135} \\

    F. W. w/o SSM Guidance
      & 0.2934
      & 0.5681
      & 1.1924
      & 1.6932
      & 2.1993 \\

    F. W. w/o Attention
      & 0.4314
      & 0.7548
      & 1.3075
      & 1.6489
      & 1.9567 \\

    w/o SSM Guidance
      & \underline{0.1663}
      & \underline{0.4241}
      & 1.2322
      & 1.7837
      & 2.3103 \\

    w/o Input Fusion
      & 0.3585
      & 0.6569
      & 0.9055
      & 1.0418
      & 1.1598 \\

    \bottomrule
  \end{tabular}
  \vspace{-4pt}
\end{table}
}

We ablate the main components of AQSS, including weighted path fusion, cross-latitude attention, attention-guided SSM modulation, and input fusion. As shown in Table~\ref{tab:ablation}, the full model achieves the lowest nRMSE at all lead times. Replacing weighted path fusion with fixed weights substantially degrades short-range accuracy, while disabling attention-guided SSM parameter modulation causes the largest long-range deterioration, increasing the 10-day nRMSE from 0.5177 to 2.3103. Removing input fusion also increases the 10-day nRMSE to 1.1598. These results indicate that weighted path fusion mainly benefits short-range prediction, while attention-guided SSM parameter modulation is critical for suppressing autoregressive error accumulation.

\section{Conclusion}
\label{sec:conclusions}

We presented an MLWP network named S$^3$N, combining L2Proj and AQSS. L2Proj is an $L^2$ finite-element mapping method, while AQSS models HP fields through attention-guided global spiral scans. Experiments show that S$^3$N achieves strong performance in long-range forecasting with lower error accumulation. Future work will focus on improving the mapping resolution between the two grids and short-range forecasting performance.

\vfill\pagebreak

\bibliographystyle{IEEEbib}
\bibliography{refs}

\end{document}